\documentclass[letterpaper]{article} 
\usepackage{aaai2026}  
\usepackage{times}  
\usepackage{helvet}  
\usepackage{courier}  
\usepackage[hyphens]{url}  
\usepackage{graphicx} 
\usepackage{natbib}  
\usepackage{caption} 
\usepackage{multirow}
\usepackage{amsmath}
\usepackage{algorithm}
\usepackage{algorithmic}

\usepackage{newfloat}
\usepackage{listings}
\DeclareCaptionStyle{ruled}{labelfont=normalfont,labelsep=colon,strut=off} 
\floatstyle{ruled}
\newfloat{listing}{tb}{lst}{}
\floatname{listing}{Listing}
\title{PersonaAgent with GraphRAG: Community-Aware
Knowledge Graphs for Personalized LLM}
\author{
    Siqi Liang\textsuperscript{\rm 1}\equalcontrib\thanks{Corresponding authors: Siqi Liang (lsq950917@gmail.com), Yudi Zhang (yudiz@iastate.edu)},
    Yudi Zhang\textsuperscript{\rm 2}\equalcontrib,
    Yue Guo\textsuperscript{\rm 3}
}
\affiliations{
    \textsuperscript{\rm 1}Purdue University\\
    \textsuperscript{\rm 2}Iowa State University\\
    \textsuperscript{\rm 3}Columbia University\\
}

\usepackage{bibentry}

\begin{document}

\maketitle

\begin{abstract}
We propose a novel framework for \emph{persona-based} language model system, motivated by the need for personalized AI agents that adapt to individual user preferences. In our approach, the agent embodies the user's ``persona" (e.g.\ user profile or taste) and is powered by a large language model (LLM). To enable the agent to leverage rich contextual information, we introduce a Knowledge-Graph-enhanced Retrieval-Augmented Generation (Graph RAG) mechanism that constructs an LLM-derived graph index of relevant documents and summarizes communities of related information. Our framework generates personalized prompts by combining: (1) a summary of the user's historical behaviors and preferences extracted from the knowledge graph, and (2) relevant global interaction patterns identified through graph-based community detection. This dynamic prompt engineering approach allows the agent to maintain consistent persona-aligned behaviors while benefiting from collective knowledge. On the LaMP benchmark, our method improves news categorization F1 by 11.1\%, movie tagging F1 by 56.1\%, and reduces product rating MAE by 10.4\% over prior methods. Our code is available at \url{https://anonymous.4open.science/r/PersonaAgentwGraphRAG-DE6F}
\end{abstract}


\section{Introduction}

Large Language Models (LLMs) have shown strong performance across applications, from recommendation tasks~\cite{xu2025,liang2025, Yu2025LLMRec, Said2025ExplainRec, Lin2023SurveyLLMRec} to agent-based systems capable of reasoning, dialogue, and tool use~\cite{Ruan2023TPTU}. While earlier work applied LLMs to isolated components of recommender systems, recent agent-based approaches address more ambitious challenges such as long-horizon decision-making, collaboration, and domain-specific reasoning~\cite{Wang2024}, often enhanced with memory, planning, retrieval, and inter-agent communication for tasks like tutoring, simulation, and assistant workflows~\cite{zou2025}.

\begin{figure*}[htbp]
\centering
\includegraphics[width=0.8\linewidth]{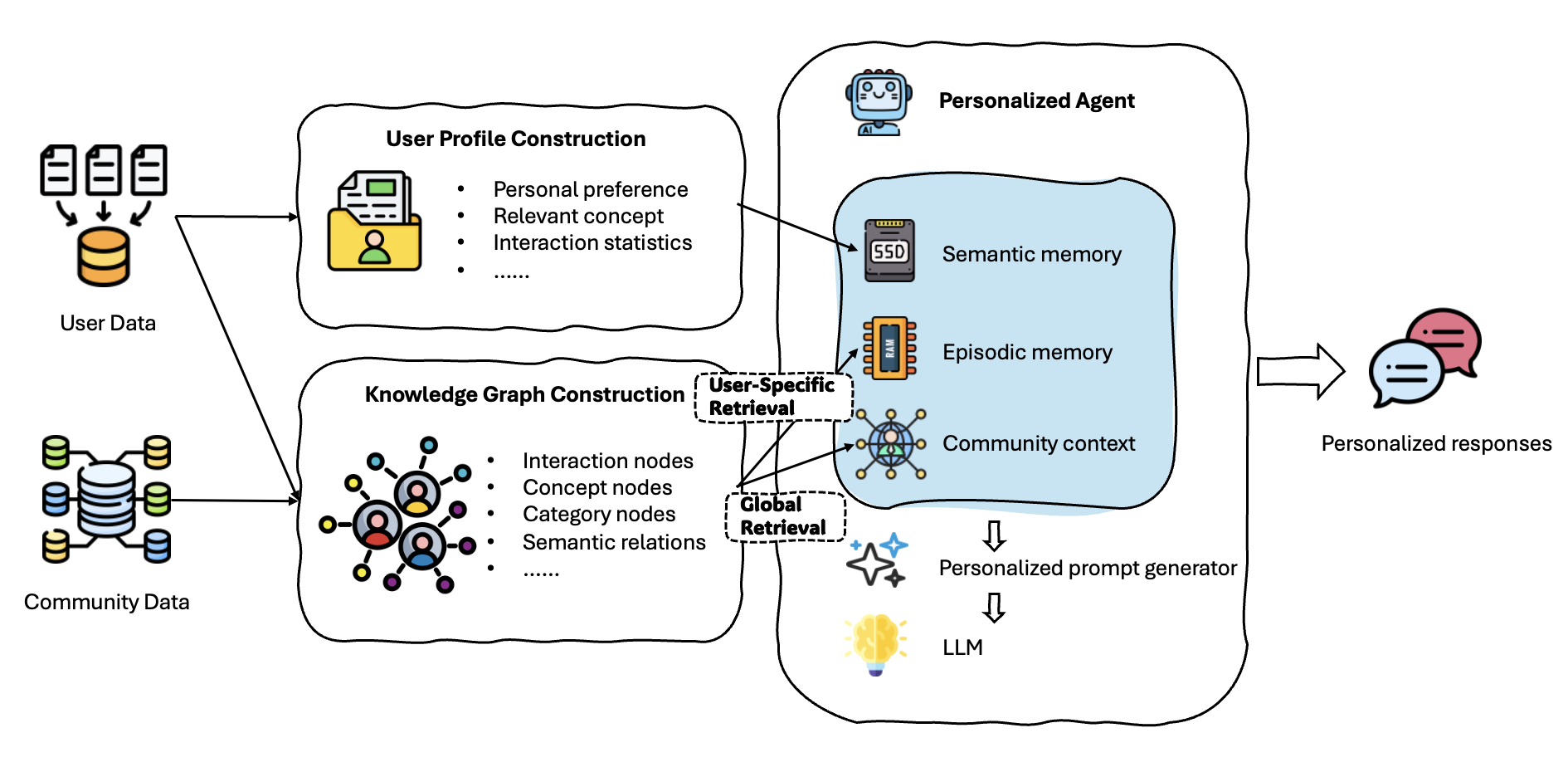}
\caption{Overview of the PersonaAgent with GraphRAG framework. }
\label{fig:personaAgent}
\end{figure*}

Within this paradigm, persona-driven agents are increasingly important for personalization: in recommender systems, an agent may reflect a user’s taste profile; in decision-support, it may simulate an expert’s reasoning style. By encoding preferences and behaviors into a natural language persona, LLMs can adapt outputs to individual users, achieving personalized behavior across dialogue, reasoning, and recommendation~\cite{harbor2024,personagym2024}. However, most prior work relies on static personas or templates, without dynamically incorporating evolving behaviors or community knowledge. In domains like movie recommendation or e-commerce, agents must produce decisions aligned with changing preferences, motivating our framework for \textit{persona-based LLM agents} that ground outputs in both individual and collective knowledge.

Our system integrates three components: (1) a \textit{persona prompt} encoding user preferences; (2) a \textit{knowledge graph} capturing personal interactions and community patterns~\cite{rpla_survey2024,mirix2024}; and (3) a \textit{GraphRAG} mechanism that retrieves and synthesizes relevant context. Given a query, dense search identifies candidate nodes, graph traversal collects related user and item signals, and the resulting subgraph is linearized and combined with the persona prompt for generation. This enables grounding in both user history and community wisdom, supporting accurate and explainable personalization~\cite{edge2024,personarag2024}.

To our knowledge, this is the first system to combine graph-based retrieval with dynamic persona prompting derived from both individual and community patterns. The result is a knowledge-aware, preference-aligned agent that improves personalization in tasks such as movie tagging and product rating.

\section{Related Work}

\subsection{Persona-Based LLM Agents}
Recent work has begun to endow LLM agents with explicit personas to achieve personalized behavior. Persona agents have demonstrated improved contextual and personalized responses across applications such as tutoring, customer support, and gaming \cite{personaagent2024}. PersonaGym~\cite{personagym2024} measures whether agents take optimal actions aligned with their personas, it assesses adherence to persona-specific communication styles, consistency in persona attributes, and avoidance of harmful outputs. 
HARBOR~\cite{harbor2024} explores how an agent's assigned persona affects its bidding behavior, whether agents can accurately profile competitors' personas during auctions.
Other studies have shown that persona prompts allow agents to extrapolate consistent preferences (e.g., adjusting answers about a tractor differently for a farmer vs. an urban planner persona). However, most prior work focuses on static personas or template-based personalization, rather than dynamically incorporating user behavior patterns and community knowledge.

\subsection{Memory and Knowledge Integration in LLM Systems} 
Memory and knowledge integration are critical for maintaining consistent and informed agent behavior. LLM-memory systems typically combine short-term context windows with long-term external memories \cite{mirix2024,rpla_survey2024}. For example, Xu et al~\cite{mirix2024} proposes a sophisticated memory architecture with multiple specialized memory types (episodic, semantic, procedural, etc) to support complex reasoning tasks. More generally, surveys have drawn analogies between human memory systems and AI memory modules \cite{rpla_survey2024}. These insights inform our approach to maintaining user preference histories and behavioral patterns.

\subsection{Retrieval-Augmented Generation and Knowledge Graphs} Retrieval-Augmented Generation (RAG) techniques use external knowledge to improve LLM outputs. Classical RAG approaches select relevant text passages via sparse term-matching or dense embedding search \cite{retrieval_survey2023,edge2024}. Recent work has extended this to graph-based knowledge structures. Graph-based RAG (GraphRAG) enriches retrieval with structured knowledge graphs: after an initial search for relevant entities, the system traverses graph links to gather related information \cite{edge2024,personarag2024}. This approach grounds LLMs in relational data, improving factual accuracy and explainability. Our framework builds on these ideas by encoding both domain knowledge and user behavior patterns in a knowledge graph, using a combination of vector retrieval and graph expansion to construct personalized contexts for the LLM.

\section{Methodology}

Our PersonaAgent system leverages Knowledge Graph-based GraphRAG to enable personalized content generation. The system combines individual user preferences with broader community insights through a structured knowledge graph and personalized prompt generation (see Fig~\ref{fig:personaAgent}).

\subsection{Knowledge Graph Construction}

Our system maintains a heterogeneous knowledge graph $G = (V, E)$ where nodes $V$ represent:
\begin{enumerate}
    \item \textbf{Interaction nodes}: Represent user interactions and contain metadata such as title, text, category, and timestamp.
    \item \textbf{Concept nodes}: Represent extracted named entities and domain-relevant keywords from interaction text. These nodes generalize across users and support semantic reasoning.
    \item \textbf{Category nodes}: Represent high-level content domains, linking interactions with broader thematic structures.
\end{enumerate}

Edges $E$ in the graph capture semantic relationships between nodes:
\begin{itemize}
    \item \textbf{Interaction $\leftrightarrow$ Category}: Connects interactions to their categorical domain.
    \item \textbf{Interaction $\leftrightarrow$ Concept}: Links an interaction to its extracted concepts or entities.
    \item \textbf{Concept $\leftrightarrow$ Concept}: Can be inferred via co-occurrence across interactions or shared categories to enable graph-based community detection.
\end{itemize}
For each new interaction, the system 1) Creates an interaction node with unique identifier; 2) Extracts relevant concepts using pattern-based and domain-specific methods;
3) Establishes connections to existing nodes based on semantic relationships.

\subsection{GraphRAG Retrieval Mechanism}

The system employs a dual-source retrieval approach that combines personal and community-based insights:

\textbf{User-Specific Retrieval} For a given user $u$ and query $q$, we retrieve relevant interactions from their history:
$$\mathcal{I}_{\text{user}}(u, q) = \text{TopK}(\text{sim}(q, i) : i \in \mathcal{H}_u)$$
where $\mathcal{H}_u$ represents user $u$'s interaction history and $\text{sim}$ refers to the Cosine similarity measured by TF-IDF.

\textbf{Global Retrieval} We augment personal context with relevant community interactions:
$$\mathcal{I}_{\text{global}}(u, q) = \text{TopK}(\text{sim}(q, i) : i \in \mathcal{H}_{\text{all}} \setminus \mathcal{H}_u)$$

The combined semantic context $\mathcal{C}(u, q)$ includes:
\begin{align}
\mathcal{C}(u, q) = \{&\mathcal{I}_{\text{user}}(u, q), \mathcal{I}_{\text{global}}(u, q), \\
&\mathcal{P}_{\text{cat}}(u), \mathcal{E}_{\text{concepts}}(u, q)\}
\end{align}
where $\mathcal{P}_{\text{cat}}(u)$ represents user category preferences and $\mathcal{E}_{\text{concepts}}(u, q)$ contains relevant concepts.

\subsection{Personalized Prompt Generation}

The system generates context-rich prompts by combining: 1) Task-specific instructions and available categories; 2) Retrieved personal interactions with relevance scores; 3) Related global community interactions and patterns; 4) User preference distributions; 5) Relevant concept clusters.

The prompt construction process follows:
\begin{algorithm}[H]
\caption{Personalized Prompt Generation}
\begin{algorithmic}[1]
\REQUIRE User ID $u$, Query $q$, Categories $\mathcal{C}$
\ENSURE Personalized prompt $P$
\STATE $context \leftarrow$ GetSemanticContext($u$, $q$)
\STATE $content \leftarrow$ ExtractTaskContent($q$)
\STATE $P \leftarrow$ InitializeBasePrompt($content$, $\mathcal{C}$)
\STATE $P \leftarrow P +$ FormatUserIntereaction
\STATE $P \leftarrow P +$ FormatCommunityIntereaction
\STATE $P \leftarrow P +$ FormatPreferencesAndConcepts($context$)
\RETURN $P$
\end{algorithmic}
\end{algorithm}
\renewcommand{\arraystretch}{1.3}
\begin{table*}[htbp]
\centering
\scalebox{1}{ 

\begin{tabular}{ccccccc}
\hline \hline
&   \textbf{Metrics}   & \textbf{Non-Personalized} & \textbf{ReAct} & \textbf{MemBank} & \textbf{PersonaAgent} & \textbf{\begin{tabular}[c]{@{}c@{}}PersonaAgent \\ with GraphRAG\end{tabular}} \\ \hline
\multirow{2}{*}{\begin{tabular}[c]{@{}c@{}}LaMP-2N: Personalized\\ News Categorization\end{tabular}} & Acc  & 0.660                     & 0.639          & 0.741            & 0.796*                 & \textbf{0.804}                                                                 \\
& F1   & 0.386                     & 0.381          & 0.456            & 0.532*                 & \textbf{0.591}                                                                 \\ \hline
\multirow{2}{*}{\begin{tabular}[c]{@{}c@{}}LaMP-2M: Personalized\\ Movie Tagging\end{tabular}}       & Acc  & 0.387                     & 0.450          & 0.470            & 0.513*                 & \textbf{0.653}                                                                 \\
& F1   & 0.302                     & 0.378          & 0.391            & 0.424                 & \textbf{0.662}                                                                 \\ \hline
\multirow{2}{*}{\begin{tabular}[c]{@{}c@{}}LaMP-3: Personalized\\ Product Rating\end{tabular}}       & MAE  & 0.295                     & 0.313          & 0.321            & 0.241*                 &   \textbf{0.216}                                                                          \\
 & RMSE & 0.590                     & 0.590          & 0.582            & 0.509*                 &    \textbf{0.484}                                                                            \\ \hline \hline
\end{tabular}
}
\caption{Performance comparison across different tasks and models}
\label{table:performance}
\end{table*}

\section{Results}

\subsection{Data Description}
We evaluate our framework using the LaMP benchmark~\cite{salemi2023lamp}, focusing on three decision-making tasks: news categorization, movie tagging, and product rating. 
These tasks enable us to assess the effectiveness of personalized agents across diverse personalization domains. Following the data processing procedure described in~\cite{personaagent2024}, we construct test sets by selecting the 100 users with the most extensive activity histories from the time-ordered version of the LaMP dataset. In the training sets which are used to construct the knowledge graph, the news data includes 274 users, the movie data includes 829 users, and product rating includes 1000 users. 



\subsection{Metircs Comparison}

Table~\ref{table:performance} presents results on three personalized tasks: news categorization (LaMP-2N), movie tagging (LaMP-2M), and product rating (LaMP-3). PersonaAgent with GraphRAG consistently outperforms all baselines including non-personalized LLMs~\cite{liu2021}, retrieval-augmented prompting (ReAct)~\cite{yao2023}, memory-based models (MemBank)~\cite{zhong2023}, and the prior state-of-the-art PersonaAgent~\cite{personaagent2024}. On LaMP-2N, it achieves 0.804 accuracy and 0.591 F1, improving over PersonaAgent by 1.0\% and 11.1\%, respectively. On LaMP-2M, the gains are larger, with accuracy increasing from 0.513 to 0.653 (+27.3\%) and F1 from 0.424 to 0.662 (+56.1\%), demonstrating stronger personalization for subjective behaviors. For LaMP-3, GraphRAG reduces MAE from 0.241 to 0.216 (–10.4\%) and RMSE from 0.509 to 0.484 (–4.9\%), indicating more precise rating prediction. 
We also noticed that with our method, small models, such as LLaMA3 can perform better than competing methods, for example, on the movie data, accuracy can be improved by 13.6\%.
These results highlight the value of integrating structured user memory with graph-based retrieval. 
\begin{figure}[htbp]
    \centering
    \includegraphics[width=\linewidth]{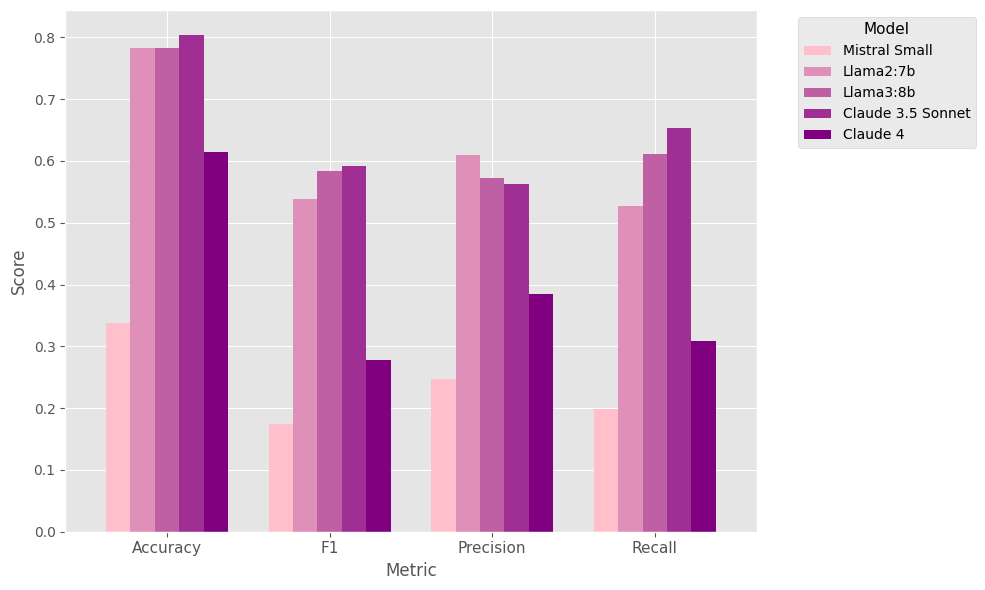}
    \caption{LLMs Comparison on LaMP-2N}
    \label{fig:llm comparison}
\end{figure}

\begin{figure*}[htbp]
    \centering
    \includegraphics[width=1\linewidth]{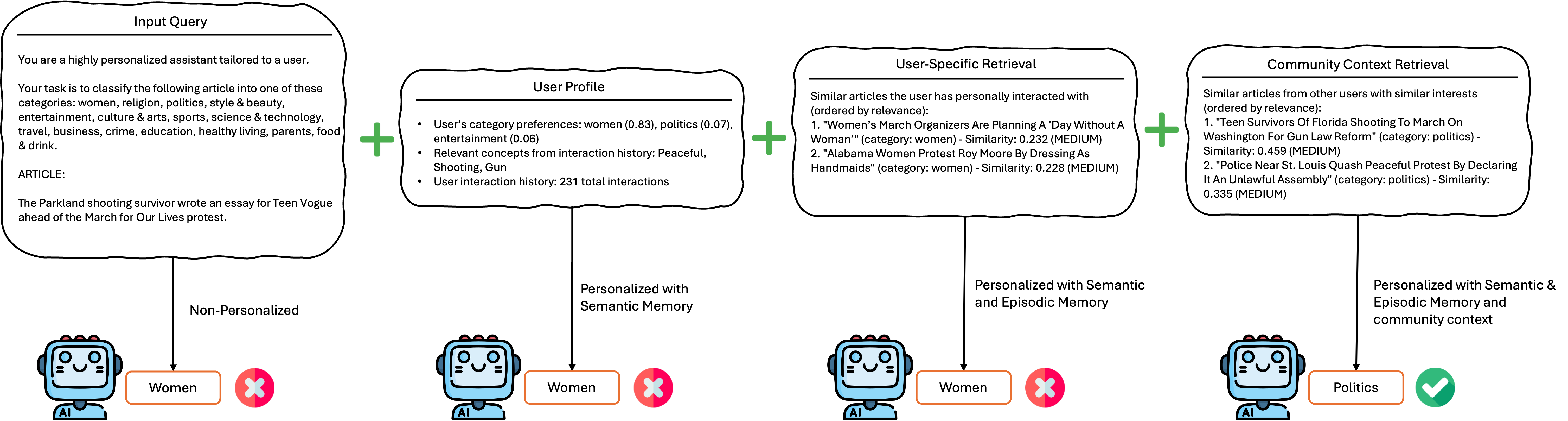}
    \caption{Case study of PersonaAgent with GraphRAG for personalized classification}
    \label{fig: case study}
\end{figure*}

Figure~\ref{fig:llm comparison} compares five language models—Mistral Small~\cite{jiang2023mistral7b}, LLaMA2-7B~\cite{touvron2023llama2}, LLaMA3-8B~\cite{llama3-8b}, Claude 3.5 Sonnet, and Claude 4~\cite{claude35}—on the LaMP-2N personalized news categorization task. Mistral Small performs worst across all metrics, reflecting its limited capacity for personalization. LLaMA2-7B shows strong results, rivaling LLaMA3-8B in Accuracy and Recall despite its smaller size, while LLaMA3-8B offers more balanced improvements in F1 and Recall. 
Among Claude models, Claude 3.5 Sonnet achieves the best overall performance, with the highest F1 and Recall, highlighting its superior alignment with user-specific content. By contrast, Claude 4 underperforms across all metrics, often overcomplicating the task and failing to provide correct answers.


\subsection{Case Study}

The example in Fig \ref{fig: case study} demonstrates that incorporating globally similar interactions from other users into the personalization prompt substantially improves classification accuracy by providing a broader contextual grounding beyond a single user’s history. In our example, the LLaMA3-8B model misclassified an article about a Parkland shooting survivor's essay for Teen Vogue as belonging to the ``women'' category when only the user’s personal interaction history was considered. This error is likely due to the user's strong historical preference for women-focused protest articles, which skewed the model's prediction. However, when we enriched the prompt with globally similar articles—such as those involving youth activism and gun law reform (e.g., ``Teen Survivors Of Florida Shooting To March On Washington'')—the model correctly classified the article as ``politics''. These globally similar interactions helped steer the model toward the correct thematic alignment by introducing relevant but more nuanced examples, thus balancing personalization with generalizability. This demonstrates that community context acts as a corrective signal, especially when a user's preferences are strongly skewed or lack diversity across topics.

\section{Conclusion}

We introduced PersonaAgent with GraphRAG, a framework that integrates persona-driven prompting with graph-enhanced retrieval to provide accurate, explainable, and consistent personalization. By leveraging both user-specific histories and global community patterns, the system balances individual preferences with collective knowledge, yielding improvements in news categorization, movie tagging, and product rating.

Looking forward, two promising directions emerge. First, multi-agent collaboration, where persona agents interact, negotiate, and share knowledge, could enhance robustness and enable collective intelligence for recommendation, classification, and decision-support. Second, incorporating inverse reinforcement learning (IRL)~\cite{beliaev2024IRLEED, swirl2024, regularizedIRL2022} would allow agents to infer latent preference signals from behavior, aligning with both explicit histories and implicit reward structures. This could produce agents that better adapt to evolving goals while remaining consistent with user values.

\appendix

\bibliography{ref}

\end{document}